\PassOptionsToPackage{backend=bibtex}{biblatex}

\documentclass{SPAICE}

\def\authorEmail{adam.abdin@centralesupelec.fr}
\def\AuthorShort{T. Leuridan et al.}

\author[]{Tristan Leuridan}
\author[]{Yi-Ping Fang}
\author[]{Adam Abdin\thanks{Corresponding author. E-Mail: \authorEmail}}
\affil[]{Department of Industrial Engineering, CentraleSupélec, Université Paris-Saclay, Gif-sur-Yvette, France}

\title{Offline Approximate Value Iteration: A Reinforcement Learning Approach for Satellite Constellation Deployment}

\begin{document}

\maketitle

\begin{abstract}
This paper introduces a reinforcement learning-based approximate dynamic programming approach to optimize the deployment of Walker-type satellite constellations under uncertainties, including launch failures and satellite malfunctions. We formulate the problem as a Markov Decision Process that incorporates low-injection orbit launches, continuous low-thrust orbit raising, and variable satellite loads. We adopt the offline approximate value iteration solution method, using linear function approximation with basis functions and action pruning strategies to substantially reduce the computational and memory requirements for finding optimal decision policies. 
Preliminary results demonstrate convergence to policies with low expected costs and deployment times, both of which are lower than those of a rolling-horizon baseline, showcasing a novel application of approximate dynamic programming in space logistics. %
\end{abstract}

\section{Introduction}

In recent years, satellite constellation deployments have surged, driven by governments and commercial entities building large-scale networks for telecommunications, navigation, and Earth observation. Mega-constellations like Starlink face major challenges, including launch failures, demand fluctuations, regulatory changes, and efficiency losses over multi-year timelines. Advances in deployment strategies and the emergence of rideshare missions have reduced entry barriers to space, yet deploying Walker-type constellations remains complex. Satellites are injected into low orbits and raised to operational altitudes via low-thrust propulsion, with rideshare injection orbits dictated by the primary payload and often requiring plane changes. This demands robust mission design, launch planning, and adaptive decision-making under uncertainty.

Existing approaches to satellite constellation deployment have relied on static optimization methods, such as the one-shot mixed-integer linear programming (MILP) on a time-expanded network \cite{sung2023optimal}, the hybrid multi-disciplinary optimization with inner mixed-integer programming \cite{di2024multi}, and integrated campaign-level space logistics formulations \cite{chen2018integrated}. 
Recent work has specifically optimized low-thrust deployment strategies by combining orbit raising and nodal drift. \cite{di2016multi,di2022optimization,pu2025simultaneously}. 
These methods all effectively solve minimum-cost/max-utility problems under perfect information. However, they cannot handle stochastic uncertainties and suffer from combinatorial explosion as the problem size grows. Sequential dynamic optimization frameworks improve adaptability, such as using a periodic static global MILP with real-time local replanning \cite{kennedy2018planning} and embedding space logistics MILP within a rolling-horizon (RH) approach for on-orbit servicing under demand uncertainty \cite{sarton2020rolling}. Related work on staged constellation deployment under demand uncertainty \cite{lee2018optimization} and reviews of space logistics modeling \cite{ho2024space} further highlight the need for methods that can adapt to evolving requirements. These methods remain myopic and computationally heavy for large fleets. Furthermore, although the constellation deployment problem has already been formulated as a Markov decision process (MDP) \cite{verfaillie2003deployment}, solving realistic instances of the problem has been limited by the curse of dimensionality. Closely related sequential decision problems in space logistics, such as autonomous on-orbit servicing under uncertainty, have also been addressed via reinforcement learning formulations of MDPs \cite{patnala2025anos} and broader operations-research frameworks for strategic OOS planning \cite{abdin2025strategic}.\\
To close these gaps (determinism and lack of foresight in static methods, and the myopic repeated solves in sequential dynamic approaches), we propose modeling the stochastic deployment as a large-scale MDP and solving it with the Offline Approximate Value Iteration (OAVI) algorithm \cite{ulmer2018budgeting,ulmer2018value,zhang2022offline}. OAVI computes a policy offline via basis-function approximation and action pruning by capturing the current constellation state, remaining satellites to deploy, and treats the uncertainties within a single MDP. This approach reduces the exponential state-action space through action pruning, and enables real-time planning for practical AI-driven constellation operations.

\section{Methodology}
\label{sec:methodology}
\subsection{Problem Description}
\begin{figure}
    \centering
    \includegraphics[width=0.66\linewidth]{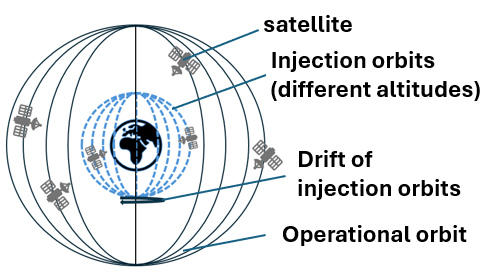}
    \caption{Constellation design (adapted from \cite{verfaillie2003deployment})}
    \label{fig:constellationdesign}
\end{figure}

\begin{figure}
    \centering
    \includegraphics[width=0.63\linewidth]{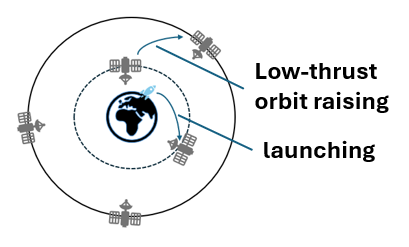}
    \caption{Satellite Actions in the constellation (adapted from \cite{verfaillie2003deployment})}
    \label{fig:deployment action}
\end{figure}

Modern large-scale constellations require deploying $N$ satellites across multiple orbital planes $P$ to reach the required target number of satellites per plane $n_{op,target}$. Yet, satellites cannot all reach their operational orbits simultaneously. They are launched in batches of up to $q_{max}$ from the ground into low altitude injection orbits with altitudes $H_{inj}$ and initial plane $p_{inj}$. To move from the injection orbits to the operational orbit, the satellite can either:
\begin{itemize}[noitemsep,leftmargin=*]
    \item remain in its injection orbit waiting to drift to the desired orbit to perform orbit raising (lower altitudes mean faster nodal precession, but higher $\Delta V$).
    \item perform a low-thrust orbit-raising maneuver to the operational altitude \(h_{op}\) while remaining in the same orbital plane.
\end{itemize}
To simplify the model, we consider only circular orbits at discrete altitudes with zero eccentricity and assume all maneuvers to be coplanar.
The constellation deployment problem considered in this study is illustrated in Figure \ref{fig:constellationdesign}, with the actions illustrated in Figure \ref{fig:deployment action}.
Low-thrust transfers to operational orbit are approximated and simplified by assuming no inclination changes, using Edelbaum's expression for the required in-plane velocity increment $\Delta V_{OR}(h_{inj})$ \cite{edelbaum1961propulsion}, followed by the integration of the Tsiolkovsky rocket equation under constant thrust to estimate maneuver duration $t_{OR}$ \cite{wertz1999space}.

We consider failures that can occur during launch, injection into orbit, or orbit-raising transfer, with respective probabilities $f_{launch}$, $f_{injection}$, and $f_{transfer}$. A satellite is operational once it is in the target orbit (no further failures are modeled).

This paper aims to find an optimal decision policy to schedule launches (timing, cadence, and quantity per launch) and plan drift/transfer decisions to deploy the exact number of operational satellites required in each orbital plane, while minimizing expected total cost, including launch, propellant, and time-dependent delay penalties.
We additionally show the framework’s ability to handle dynamic environmental changes. Specifically, updates of the target constellation to a new desired configuration (potentially with a higher or lower number of satellites) during the long-term deployment horizon. This feature provides a realistic representation of scenarios such as regulatory changes, evolving mission requirements, and fluctuations in service demand.

\subsection{The Markov Decision Process Model}
We formulate the constellation deployment problem as a finite-horizon MDP. Each time-step $t$ is a full decision cycle triggered by the completion of previous cycle's launch scheduling

At $t=0$, all orbital planes $p\in P$ are empty ($n_{inj}=\mathbf{0}$, $n_{raising}=\mathbf{0}$, $n_{op}=\mathbf{0}$) and no launch is scheduled ($ptl=\mathbf{0}$). The episode ends at \(t=T\) when the target constellation is fully deployed \(\forall p: n_{op}[p] \geq n_{op,target}[p]\). Note that \(T\) is realization-dependent but bounded.
Each time-step \(t\) consists of three sequential phases \(\phi = 0,1,2\). Each phase has a decision state \(s_t^\phi\) and for phase $\phi=2$, there is a corresponding post-decision state which is defined for the solution method presented in subsection \ref{subsec:algorithmimp}. The state is fully observable. Phase-specific actions determine the launch parameters in phase $\phi=0$, the transfers to operational orbits in phase $\phi=1$, and the planning of future launches in phase $\phi=2$.

The launch pipeline \(ptl\) tracks planned launches over the decision horizon \(DH\)(the maximum number of time-steps into the future for which launch decisions can be planned)), where each \(ptl_t[i] \in \{0,1\}\) is a binary indicator of whether a launch is scheduled \(i\) time-steps ahead (\(ptl_t[0]\) for the current time-step).

Each decision increases the total expected cost and contributes to the deployment duration. Our goal is to find the optimal policy \(\pi\) that minimizes the expected total cost while also reducing the duration by adding delay penalties.

\subsubsection{State Description}
At each time-step $t$, the constellation state includes the current status of satellites across injection orbits, orbit-raising trajectories, and operational orbits, as well as the pending launch pipeline.

We therefore define the state $s^\phi_t$ at time $t$ and phase $\phi\in \{0,1,2\}$ as a tuple comprising the following elements:
\begin{itemize}[noitemsep,leftmargin=*]
    \item for each orbital plane $p = \{1, \dots, P\}$ and each injection altitude $h =\{1,\dots,H_{inj}\}$, the number $n_{inj,t}[p,h] \geq 0$ of satellites currently in the injection orbit at altitude $h$ in plane $p$;
    \item for each orbital plane $p$, the number $n_{op,t}[p] \geq 0$ of operational satellites already in the operational orbit at altitude $h_{op}$ in plane $p$;
    \item for each orbital plane $p$ and each remaining delay index $d\in \{0,1,...,d_{\max} \}$, the number $n_{\text{raising},t}[p,d] \geq 0$ of satellites currently in low-thrust orbit-raising transfer from plane $p$ with $d$ time-steps remaining until arrival in the operational orbit;
    \item the current launch schedule $ptl_t$;
    \item and the target number of operational satellites in each planes $n_{op,target}$.
\end{itemize}

Thus, the full state is represented as\\ $s^\phi_t= \bigl( {n_{inj,t},\ n_{op,t},\ n_{\text{raising},t},\ ptl_t,n_{op,target}} \bigr)$.

\subsubsection{Decisions}
At each decision epoch $t$, the agent makes three sequential phase-dependent decisions:
\begin{itemize}[noitemsep,leftmargin=*]
    \item In phase $\phi=0$ (launch commitment): choice of the target plane $p$, the injection altitude  $h$, and the number of satellites to launch $q = \{0,1,\dots, q_{\max}\}$ subject to the launcher capacity $q_{max}$. Launches are constrained by the remaining satellite needs. If $ptl_t[0]=0$ or the constellation is already complete, only the no-commit action $(0,0,0)$ is allowed.
    \item In phase $\phi=1$ (orbit-raising initiation): for each plane $p$, the agent chooses how many satellites to transfer from injection to operational orbit $n_{OR}[p,\cdot] \in \{n_{inj}[p,\cdot]\}$, if any at all. The choice is limited by the available satellites in $n_{inj}[p,\cdot]$ and by the projected shortage in the operational orbit (considering incoming $n_{\text{raising}}[p,\cdot]$).
    \item In phase $\phi=2$ (next launcher scheduling): Decision about whether or not to schedule a new launch $ptl[DH] \in \{0, 1\}$. This decision updates the schedule array $ptl$ and inserts the new decision at DH time-steps, ensuring $ptl$ is consistent when phase 0 of the next time-step begins. Finally, the immediate costs are applied.
\end{itemize}

The full action $a_t$ is the concatenation of the three-phase decisions executed sequentially within time-step $t$. The action space is therefore phase-dependent and pruned dynamically to avoid infeasible or redundant choices \cite{zhang2022offline}.

\subsubsection{State Transitions}
State transitions combine deterministic phase execution and stochastic updates due to launch, injection, and transfer failures. %
Launch failures are accounted for in phase 0, while injection and transfer failures are accounted for in phase 2. %
 Actions in $\phi=0$ inject either all or no satellites to $n_{inj}$ with success probability $1-f_{launch}$ while actions in $\phi=1$ move, deterministically, satellites from $n_{inj}$ to $n_{\text{raising}}$ with delay $d$ based on low-thrust duration and drift updates across planes of $n_{inj}[:,h]$ layers. In Phase~2, for each step, satellites in the injection orbit \(n_{inj}[p,\cdot]\) fail with probability \(f_{injection}\), while raising satellites \(n_{raising}[p,0]\) fail with probability \(f_{transfer}\) and are lost if they do so; successful raising maneuvers are added to the operational orbit \(n_{op}[p]\), and the launch-decision pipeline is shifted: $ptl_t[i] \leftarrow ptl_t[i+1]$; $ptl_t[DH] \leftarrow \delta_t$ where $\delta_t\in\{0,1\}$ is the new binary decision for launch.

\subsubsection{Objective Function}
The objective is to find a policy $\pi$ that minimizes the expected total discounted cost over the finite deployment horizon as in Eq. \ref{eq:pimin}
\begin{equation}
    \min_{\pi} \ \mathbb{E} \left[ \sum_{t=0}^{T} \gamma^t \, c_t(s_t, \pi(s_t), s_{t+1}) \right],
    \label{eq:pimin}
\end{equation}
where $\gamma \in [0,1]$ is a discount factor. The immediate cost $c_t(s_t, a_t, s_{t+1})$  realized in phase 2, is described in Eq. \ref{eq:cost_function}:
\small
\begin{equation}
\begin{aligned}
c_t &= 
  C_L \mathbf{1}\{q_t>0\} + C_S \cdot q_t+\sum_{p=1}^{P} C_U(\text{m}_p)\\ &+ n_{OR} \cdot \Delta V_{OR}(h_{inj}) \cdot c_{fuel}- n_{dep} \cdot R_{dep} \\&-R_{\text{end}} \mathbf{1}\{\forall p: n_{op}[p]\geq n_{op,target}[p]\}
  \end{aligned}
  \label{eq:cost_function}
\end{equation}
where $C_L$ and $C_S$ are launcher fixed cost and per-satellite costs, $C_U(m)$ is the unavailability cost for plane $p$ missing $m$ satellites, $n_{OR}$ is the number of satellites initiating an orbit raising maneuver in that timestep,  $\Delta V_{OR}(h_{inj})$ is the $\Delta V$ needed for a satellite to maneuver from injection altitude $h_{inj}$ to operation altitude $h_{op}$, $c_{fuel}$ is the fuel cost per $\Delta V$ used, $n_{dep}$ is the number of satellites successfully deployed, and $R_{dep}$ is the reward per successfully deployed satellites. A large positive terminal reward $R_{\text{end}}$ is added upon successful completion ($\forall p:\ n_{op} \geq n_{op,target}$).

\subsection{Algorithm Implementation}
\label{subsec:algorithmimp}
Due to the high dimensionality of the state and action spaces, exact value iteration is intractable. We therefore apply OAVI with linear value function approximation (VFA) and action-space pruning to obtain high-quality approximate policies after offline training. The overall methodology belongs to the family of approximate dynamic programming and reinforcement learning techniques for large-scale sequential decision problems under uncertainty \cite{powell2011approximate,sutton2018reinforcement,bertsekas2012dynamic}.

The VFA is learned offline using post-decision states (PDS), which are defined as the system state immediately after the deterministic effects of the chosen action in phase 2 but before the stochastic transitions realize. PDS effectively decouple deterministic decision effects from stochastic transitions. This separation enables efficient computation of expected future values without re-simulating uncertainties at each step, allowing for fast online greedy action selection.

Training proceeds via iterative Monte Carlo rollouts over multiple episodes. At each decision epoch, candidate actions are generated and evaluated using the current VFA on the resulting post-decision states. With probability $1 - \epsilon$ (where $\epsilon$ decays over iterations), the action yielding the highest estimated Q-value is selected greedily; otherwise, a random feasible action is chosen. After executing the action, the immediate reward and the PDS feature vector are used to compute the temporal-difference target. The linear parameters $\theta$ of the VFA are then updated via a classical offline least-squares approach on accumulated samples, iterating until convergence.

To further improve the computational performance, we apply two pruning strategies: recourse reduction, which prevents transfers to saturated planes ($n_{op}\geq n_{op,target}$), and neighborhood reduction, which biases launch actions toward planes with the largest satellite shortages. These strategies are enforced in our model, significantly reducing the action state space compared to unpruned space.

\begin{figure}[h!]
    \centering
    \includegraphics[width=0.7\linewidth]{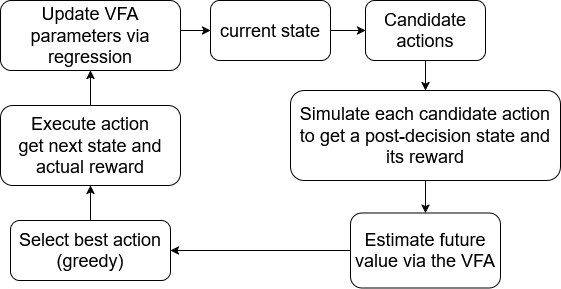}
    \caption{OAVI implementation flow diagram}
    \label{fig:OAVI}
\end{figure}

We approximate the value function with a 28-dimensional linear feature set. Features include global deployment and launch indicators, constellation-level aggregates of operational shortage, effective shortage (target minus current operational satellites minus orbit raising satellites), and operational count. This design targets the quantities that govern launch timing and residual deployable demand. Although more expressive features—such as a priori cost and credibility terms—have proven effective in related multivehicle routing settings~\cite{zhang2022offline}, we observed that they degraded generalization in our offline regime with limited samples. A conceptual overview of the OAVI procedure is provided in Figure~\ref{fig:OAVI}.

\section{Results and Analysis}
\label{sec:Results}
For initial results, we evaluate two 4-orbital-plane constellations starting with a target of 4 operational satellites per plane.
One keeps the original target; the other is updated at timestep $10$ to include 2 additional satellites in plane 1 and 1 in plane 3.
The time-step is $0.5$ months, and the launcher has a total launch capacity of $q_{max}=2$.
We consider the configurations with the specified decision horizon, failure probabilities, costs in $\text{M}\$$, orbital and injection heights, and propulsion parameters as seen in Table \ref{tab:params}.
The following rewards are used, $R_{\text{dep}}$= 20, $R_{\text{end}}=200. $%
\\
To assess the scalability of the proposed approach beyond the initial small-scale setting, we additionally evaluate a larger constellation comprising 15 orbital planes, with a target of 10 operational satellites per plane. Furthermore, the launcher has an updated total launch capacity of $q_{max}=10$.  For this larger instance, the reward parameters were scaled to $R_{\rm dep}=40$ and $R_{\rm end}=400$. All other model parameters remain identical to those of the small-scale scenario (Table \ref{tab:params}) except for the unavailability cost $C_U$ which is kept identical up to 4 satellites and then adds 3 per additional missing satellite so full-plane outages scale with fleet size.

Two rolling-horizon baselines were implemented under identical stochastic parameters. At each decision point, a deterministic MILP is solved over a receding window of H=24 steps (12 months), and the problem is re-solved on the realized state. The simplified rolling horizon algorithm does not use the 3-phase structure and omits passive inter-plane drift, yielding a leaner model in which the optimizer faces fewer structural constraints. For a fairer comparison, the second rolling-horizon algorithm is embedded in the same 3-phase simulator as the OAVI algorithm, with replanning at the start of each month (phase 0). Both variants were evaluated over 50 Monte Carlo rollouts for all scenarios. Shorter lookaheads (H=6, 3 months) fail to complete the mission, confirming that larger horizon are required. %
\begin{table}[htbp]
\centering
\small
\renewcommand{\arraystretch}{0.75}
\begin{tabular}{@{}ll@{}}
\toprule
\textbf{Parameter} & \textbf{Value} \\
\midrule
$DH$ & 4 \\
$d_{\max}$ & 7 \\
$\gamma$ & 0.99 \\
$q_{\max}$ & 2 \\
$f_{\rm launch}$ & 0.1 \\
$f_{\rm injection}$ & 0.01 \\
$f_{\rm transfer}$ & 0.005 \\
$C_L$ & 10 \\
$C_S$ & 0.7 \\
$C_U$ & [0, 4, 6, 9, 15] \\
$c_{\rm fuel}$ & 0.0001 \$M/m/s \\
Injection heights & [250, 450]\,km \\
Drift Rates & [0.5 months, 1 months] \\
Operational height & 1000\,km \\
$I_{\rm sp}$ & 3000\,s \\
$m_0$ & 350\,kg \\
Thrust & 0.02\,N \\
\bottomrule
\end{tabular}
\vspace{-0.6em}         
\caption{Model parameters}
\label{tab:params}
\end{table}
\begin{figure}[!t]
    \centering
    \includegraphics[width=0.9\linewidth]{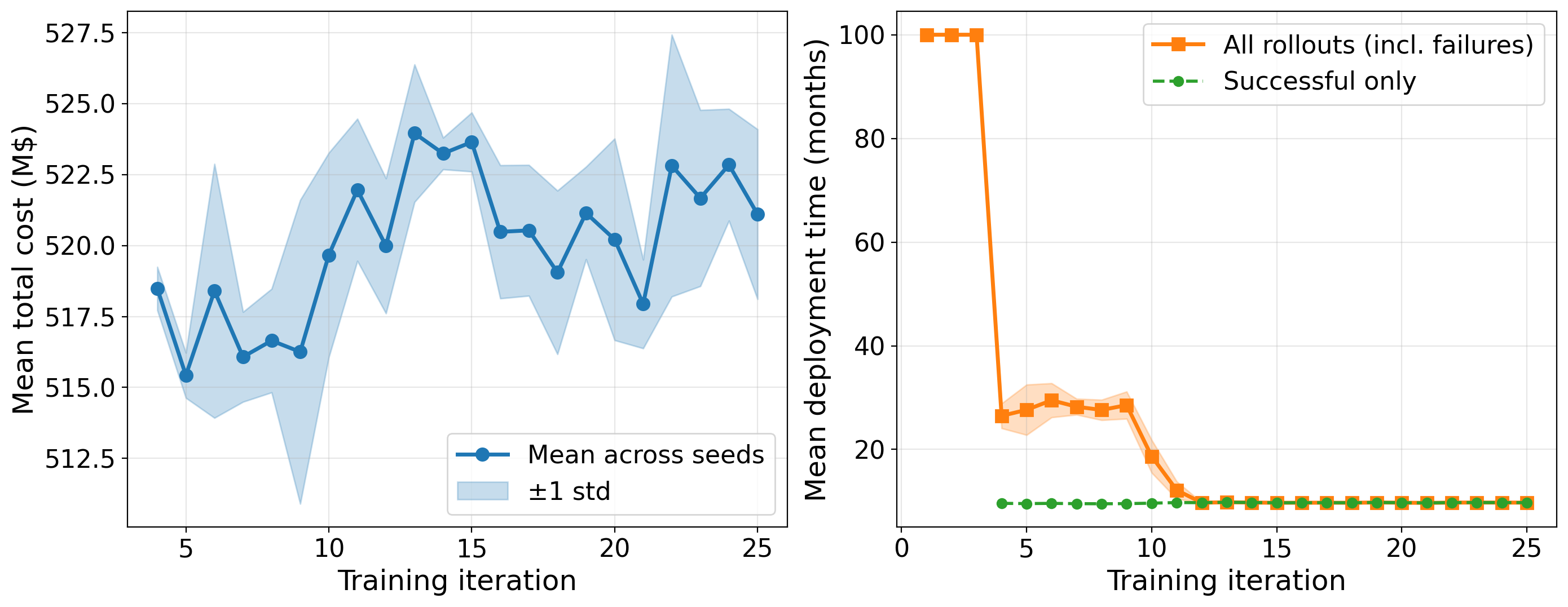}
    
    \vspace{0.8em}
    
    \includegraphics[width=0.9\linewidth]{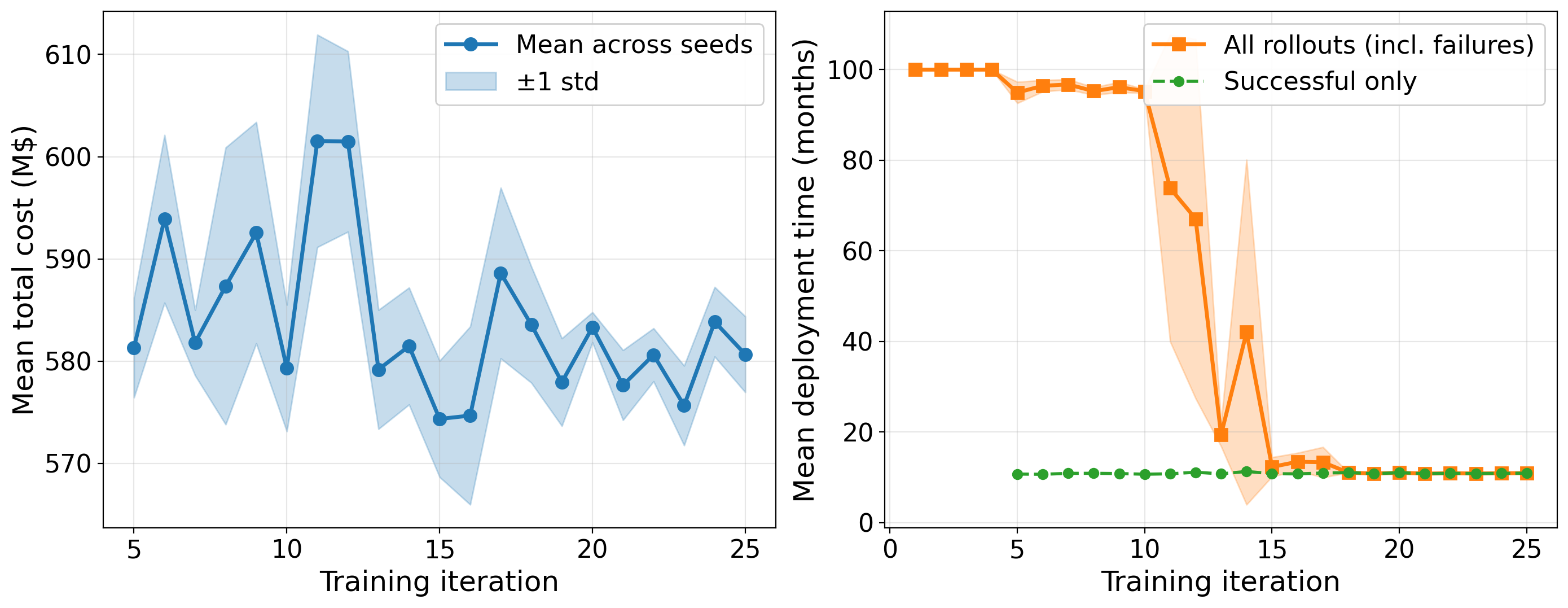}

    \includegraphics[width=0.9\linewidth]{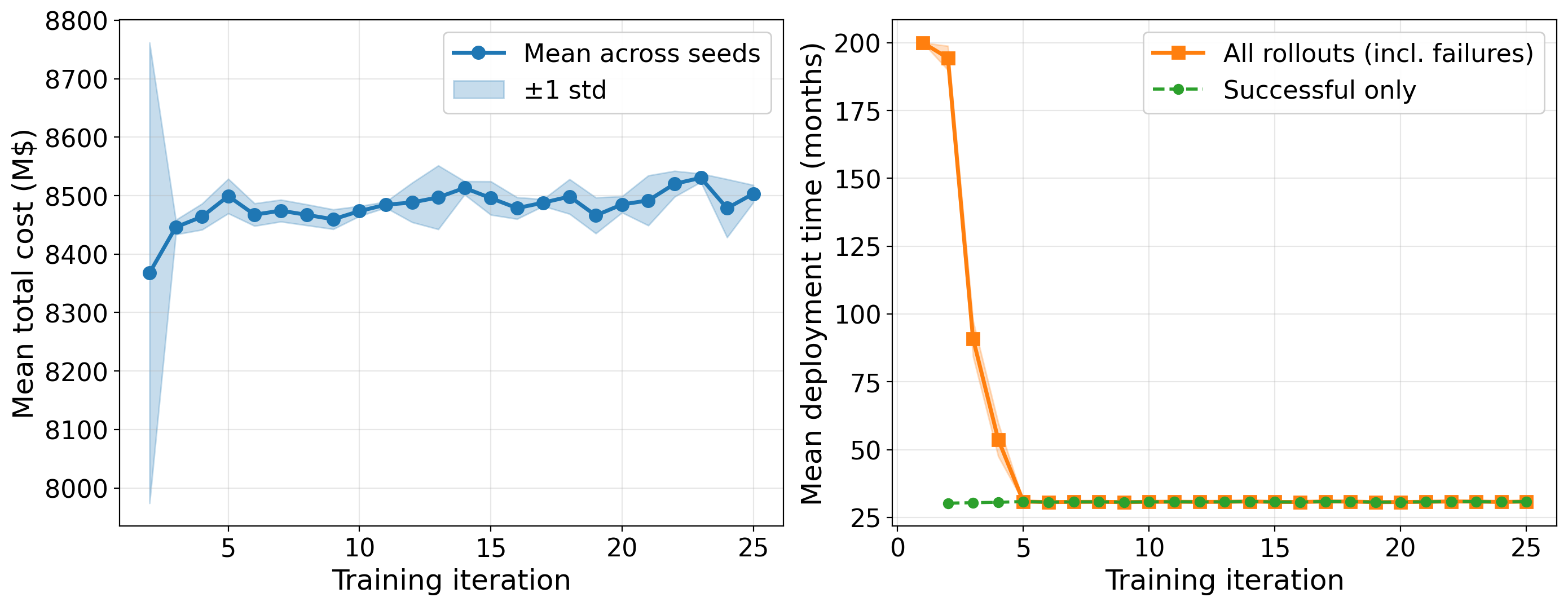}
    
    \caption{Top: standard iteration learning. Middle: iteration learning with constellation update convergence results.
    Bottom: iteration learning for large constellation deployment}
    \label{fig:convergence-combined}
\end{figure}

Training is performed over 25 iterations, with 100 episodes per iteration. Figure \ref{fig:convergence-combined} shows the evolution of mean total cost and mean deployment time over iterations (averaged over 100 stochastic rollouts). Across all three constellation sizes, the cost curves remain largely flat or exhibit only mild oscillations. Likewise, the deployment time of successful episodes is essentially constant from the first iteration in which the constellation is successfully deployed. The primary improvement observed during training is a progressive reduction in the deployment time for all episodes. This indicates that the learning process primarily increases the success rate across all rollouts. The agent becomes more reliable at finding feasible deployments.

\begin{table}[h!]
        \centering
        \footnotesize
        \begin{tabular}{llccc}
            \toprule
            \textbf{Scenario} & \textbf{Method} & \textbf{Duration (mo)} & \textbf{Total Cost (M\$)} \\
            \midrule
            Normal &OAVI & 9.7 $\pm$ 0.9 & 521.1 $\pm$ 32\\
           & RH (lite) & 9.0 $\pm$ 0.5 & 485.2 $\pm$ 28\\
           & RH & 10.6 $\pm$ 1.2 & 555.6 $\pm$ 46 \\
            \midrule
            Month-5 update & OAVI &10.9 $\pm$ 0.3 &580.7 $\pm$ 37\\
            & RH (lite) & 10.0$\pm$ 0.5 & 541.5 $\pm$ 29\\
            & RH & 13.1 $\pm$ 1.7 & 677.1$\pm$ 76.5\\
            \midrule
            Large constellation & OAVI &30.8 $\pm$ 1.4 &8492.6 $\pm$371.3\\
            & RH (lite) & 25.9 $\pm$ 2.4& 8005.2 $\pm$ 346.2\\
            & RH & 40.6 $\pm$ 2.4& 10950.9 $\pm$ 347.2 \\
            \bottomrule
        \end{tabular}
        \vspace{-0.8em}    
        \caption{Algorithm Performance}
        \label{tab:algosol}
    \end{table}
    
Table \ref{tab:algosol} summarizes the performance of the three algorithms. Our solution converges to policies achieving a mean deployment time of $  9.7  $ months and a cost of $  \$521.1  $ M in the no-update scenario, $  10.9  $ months, $  \$580.7  $ M in the updated constellation scenario, and  $30.8$ months, $  \$8492.6  $ M for the large constellation deployment. It outperforms the traditional rolling-horizon method, particularly in the updated constellation and large constellation scenario, where the baseline exhibits substantially larger deviations from the optimal solution in both deployment time and cost.

\begin{figure}[!t]
    \centering
    \includegraphics[width=.9\linewidth]{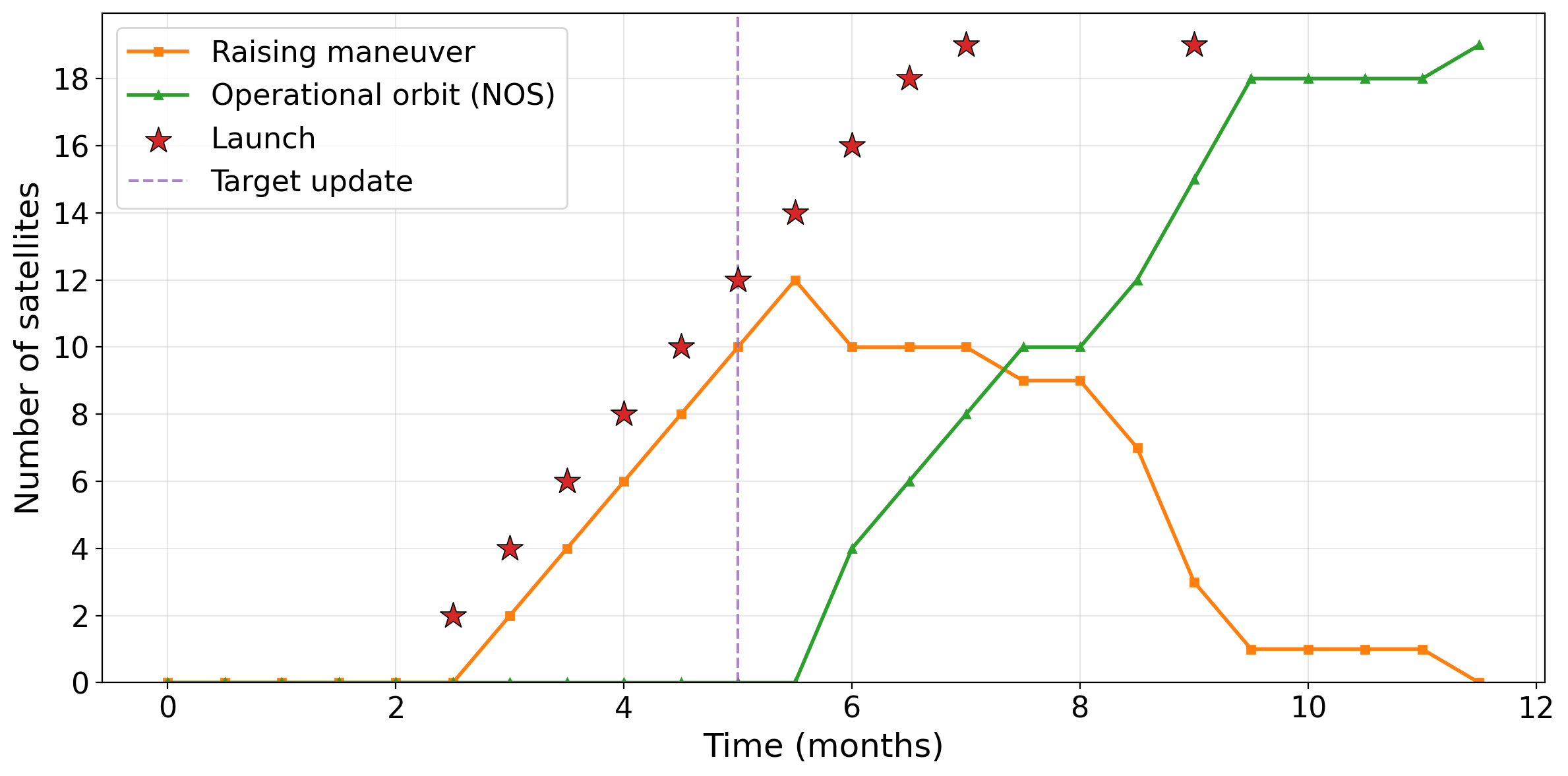}
	\caption{Deployment process}
	\label{fig:deployment timeline} 
\end{figure}

Figure \ref{fig:deployment timeline} illustrates a representative episode under the converged policy ($  \epsilon=0  $).
Satellites are maneuvered directly from injection into their operational orbits, avoiding inefficient planes and long drift periods, thereby shortening deployment time.
We observe that the deployment proceeds continuously with no overdeployment until the final launch fails. The algorithm must then schedule a replacement launch, which delays completion and increases the overall deployment time. Furthermore, by deploying the single satellite last, the algorithm effectively minimizes the unavailability cost.

\begin{table}[h!]
\centering
\small
\setlength{\tabcolsep}{3pt}
\begin{tabular}{lrrrrrrrr}
\toprule
\textbf{months} & \textbf{6.0} & \textbf{6.5} & \textbf{7.0} & \textbf{7.5} & \textbf{8.5} & \textbf{9.0} & \textbf{9.5} & \textbf{11.5} \\
\midrule
Plane 1 & 0 & 0 & 0 & 2 & 2 & 3 & 5 & \textbf{6} \\
Plane 2 & 2 & 2 & \textbf{4} & \textbf{4} & \textbf{4} & \textbf{4} & \textbf{4} & \textbf{4} \\
Plane 3 & 2 & 4 & 4 & 4 & 4 & 4 & \textbf{5} & \textbf{5} \\
Plane 4 & 0 & 0 & 0 & 0 & 2 & \textbf{4} & \textbf{4} & \textbf{4} \\
\midrule
\textbf{Total} & 4 & 6 & 8 & 10 & 12 & 15 & 18 & \textbf{19} \\
\bottomrule
\end{tabular}
\caption{Operational satellites per plane during evolving-plan deployment example}
\label{tab:operational-satellites}
\end{table}

\begin{figure}
    \centering
    \includegraphics[width=\linewidth]{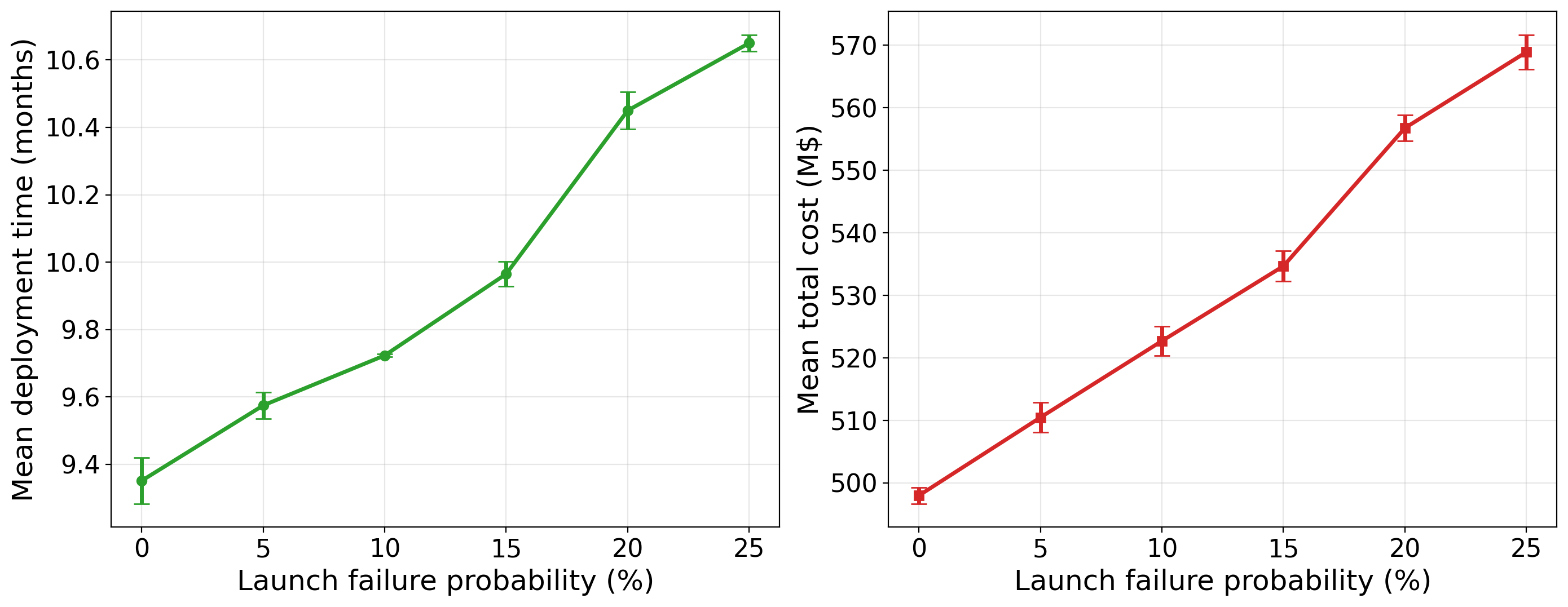}
    \caption{Sensitivity to launch failure for cost and deployment of the constellation }
    \label{fig:sensanalysis}
\end{figure}

Table \ref{tab:operational-satellites} displays the number of operational satellites per orbital plane for the months with updates in the operational plane.
As shown in the table, the policy correctly handles the update introduced at month 5. Plane 1 gains its 2 additional satellites after 11.5 months, while plane 3 achieves its revised target of 5 satellites by month 9.5. The remaining planes follow the original design. Overall, the policy effectively incorporates the required 3 new satellites during deployment, resulting in a total of 19 operational satellites by month 11.5.
\\
Finally, a sensitivity analysis was conducted on launch failure probability for the no-update scenario, as this parameter most strongly affects deployment time and cost by requiring replanning after a failed launch and resulting in the loss of all satellites. As shown in Figure \ref{fig:sensanalysis}, there is a clear impact of the failure probability on both cost and deployment time, both of which increase gradually with rising failure probability.

\section{Conclusion}
\label{sec:Conclusion}
This paper addresses the dynamic policy optimization of Walker-type satellite constellation deployment under uncertainties, including launch failures, satellite malfunctions, and orbit-raising failures. We formulate the problem as a finite-horizon MDP, incorporating low-injection orbit launches, continuous low-thrust orbit raising, and variable satellite loads to minimize expected costs (launch, propellant, and delay penalties). To reduce computational and memory requirements for policy training in this large-scale MDP, we adapt the RL-based OAVI algorithm, using linear VFA with basis functions and action pruning strategies.
Preliminary simulation results confirm that the adapted OAVI successfully solves the multi-plane deployment problem with high sample efficiency, fast online evaluation, and  handling of dynamic environmental changes. 
The algorithm converges to policies that achieve low expected costs and short deployment times. It outperforms traditional rolling-horizon baselines, particularly for updated constellations, demonstrating reinforcement learning as a promising approach for approximate dynamic programming in AI-driven space logistics. While the present study has focused primarily on smaller constellations to establish and validate the proposed algorithm, the method was also applied to a larger constellation configuration. In that case, convergence to a solution was demonstrated, and a feasible solution was compared against baseline approaches; however, a more extensive performance analysis was not carried out. Future research could therefore extend the evaluation to still larger configurations, enabling a rigorous stress-test of the algorithm, clarification of its scalability limits, and a clearer assessment of its practical suitability for real-world mega-constellation scenarios.

\begin{acknowledgments}

\texttt{This work was supported by the French National Research Agency (ANR) under grant \textbf{ANR-23-CMAS-0041}, Académie Spatiale d'Île-de-France
project\footnote{https://academiespatiale.fr/}, grant \textbf{ANR-23-IACL-0003} (DATAIA Cluster, France 2030 IA Cluster programme), and grant \textbf{ANR-23-CE10-0006} (Resilient and Sustainable Planning and Management of Future Space Industry Infrastructure with On-Orbit Servicing). This research was also supported by the European Union's Horizon Europe research and innovation programme under the Marie Skłodowska-Curie COFUND grant agreement No.\ 101127936 (DeMythif.AI).}

\end{acknowledgments}

\printbibliography
\addcontentsline{toc}{section}{References}

\end{document}